\documentclass[runningheads]{llncs}
\usepackage{graphicx}
\usepackage{hyperref}
\usepackage{color}

\usepackage[binary-units=true,detect-weight=true]{siunitx}

\usepackage{cite}

\usepackage{amsmath}
\usepackage{amsfonts}

\usepackage{todonotes}

\newcommand{\spass}{SPASS}
\newcommand{\vampire}{Vampire}

\newcommand{\smtlib}{SMT-LIB}

\title{Vampire With a Brain Is a Good ITP Hammer}
\author{Martin Suda
\orcidID{0000-0003-0989-5800}}
\authorrunning{M. Suda}
\institute{Czech Technical University in Prague, Czech Republic \\
\email{martin.suda@cvut.cz}}

\begin{document}

\maketitle

\begin{abstract}
Vampire has been for a long time the strongest first-order automatic
theorem prover, widely used for hammer-style proof automation in ITPs
such as Mizar, Isabelle, HOL, and Coq. In this work, we considerably
improve the performance of Vampire in hammering over the full Mizar
library by enhancing its saturation procedure with efficient neural
guidance.  In particular, we employ a recently proposed recursive neural network
classifying the generated clauses based only on their derivation
history.  Compared to previous neural methods based on considering the
logical content of the clauses, our architecture makes evaluating a single clause
much less time consuming. The resulting system shows good learning
capability and improves on the state-of-the-art performance on the Mizar
library, while proving many theorems that the related ENIGMA system
could not prove in a similar hammering evaluation.
\end{abstract}


\section{Introduction}

The usability of interactive theorem provers (ITPs) is significantly enhanced by proof automation.
In particular, employing so-called \emph{hammers} \cite{DBLP:journals/jfrea/BlanchetteKPU16}, systems that connect the ITP to 
an automatic theorem prover (ATP), may greatly speed up the formalisation process.

There are two ingredients of the hammer technology that appear to be best implemented 
using machine learning, especially while taking advantage of the corresponding large ambient ITP libraries,
which can be used for training. One is the \emph{premise selection} task,
in which the system decides on a manageable subset of the most relevant facts from the ITP library
to be passed to the ATP as axioms along with the current conjecture 
\cite{DBLP:journals/jar/AlamaHKTU14,DBLP:conf/frocos/FarberK15,DBLP:journals/corr/AlemiCISU16,DBLP:conf/nips/WangTWD17,DBLP:conf/lpar/PiotrowskiU20}.
The other is the \emph{internal guidance} of the ATP's proof search \cite{DBLP:conf/tableaux/UrbanVS11,DBLP:conf/cade/FarberKU17}, 
where a machine-learned component 
helps to resolve some form of don't-care non-determinism in the prover algorithm with the aim of speeding up the proof search.
In the predominant saturation-based proving paradigm, employed by the leading ATPs such as E~\cite{SCV:CADE-2019}, \spass{} \cite{DBLP:conf/cade/WeidenbachDFKSW09}, or \vampire{}~\cite{DBLP:conf/cav/KovacsV13},
internal guidance typically focuses on the \emph{clause selection} choice point \cite{DBLP:conf/mkm/JakubuvU17,DBLP:conf/lpar/LoosISK17}.

ENIGMA \cite{DBLP:conf/mkm/JakubuvU17,DBLP:conf/mkm/JakubuvU18,DBLP:conf/cade/ChvalovskyJ0U19,DBLP:conf/cade/JakubuvCOP0U20} is a system
delivering 
internal proof search guidance driven by state-of-the-art machine learning methods to the automatic theorem prover E~\cite{SCV:CADE-2019}.
In 2019, the authors of ENIGMA announced \cite{DBLP:conf/itp/JakubuvU19} a \SI{70}{\percent} improvement 
(in terms of the number of problems solved under a certain wall clock time limit)
of E on the Mizar mathematical library (MML) \cite{DBLP:journals/jfrea/GrabowskiKN10}.
This was achieved using gradient boosted trees coupled with a clause representation by efficiently extracted manually designed features.



In our recent work \cite{SudaCADE2021}, we presented an enhancement of the automatic theorem prover \vampire{} \cite{DBLP:conf/cav/KovacsV13}
by a new form of clause selection guidance. The idea is to employ a recursive neural network \cite{DBLP:conf/icnn/GollerK96} and 
to train it to classify clauses based solely on their derivation history. This means we deliberately abstract away the logical content of a clause,
i.e.~``what the clause says'', and only focus on ``where the clause is coming from (and how)''.  %
There is a pragmatic appeal in this design decision: evaluating a clause becomes relatively fast compared 
to other approaches based on neural networks (cf., e.g., \cite{DBLP:conf/lpar/LoosISK17,DBLP:conf/cade/ChvalovskyJ0U19}).
It is also very interesting that such a simple approach works at all,
let alone being able to match or even improve on the existing ``better informed'' methods.


We originally developed and evaluated \cite{SudaCADE2021} the architecture in the experimental setting of theory reasoning over the \smtlib{} library.
In this paper, we instead explore its utility for improving the performance of \vampire{} in the role of an ITP hammer,
focusing on the well-established Mizar benchmark \cite{DBLP:journals/jar/KaliszykU15a}.
Mizar requires the architecture to be adapted to a different set of features, notably a much larger set of background axioms (used instead of theory axioms) and a conjecture.
While we previously \cite{SudaCADE2021} evaluated various modes of integrating the learned guidance and some supporting techniques,
here we conduct new experiments that shed light on how the behaviour of the prover changes with varying parameters of the network itself.
Finally, our evaluation allows for a direct comparison with the ENIGMA work of Jakub\r{u}v and Urban \cite{DBLP:conf/itp/JakubuvU19}.

In the rest of this paper, we first recall (in Sect.~\ref{sec:enigma_style}) how the saturation-based ATP technology can be 
enhanced by internal guidance learnt from previous proofs. 
We then explain (in Sect.~\ref{sec:neural})
how to construct and train recursive neural networks as successful classifiers of clause derivations.
While there is a certain overlap with our previous work \cite{SudaCADE2021}, the new benchmark
allows for the incorporation of conjecture-related features (Sect.~\ref{subsect:levels})
and we also find room to explain how to efficiently train our networks using parallelisation (Sect.~\ref{subsect:train_in_par}).
Finally, we report (in Sect.~\ref{sec:exper}) on an experimental evaluation  of our extension of \vampire{} with the described techniques
over the Mizar mathematical library.




\section{Internal Guidance of an ATP using Machine Learning}

\label{sec:enigma_style}

Modern automatic theorem provers (ATPs) for first-order logic 
such as E~\cite{SCV:CADE-2019}, \spass{} \cite{DBLP:conf/cade/WeidenbachDFKSW09}, or \vampire{}~\cite{DBLP:conf/cav/KovacsV13} 
are one of the most mature tools for general reasoning in a variety of domains.
In a nutshell, they work in the following way.

Given a list of \emph{axioms} $A_1,\ldots,A_l$ and a \emph{conjecture} $G$ to prove,
an ATP translates $\{A_1,\ldots,A_l,\neg G\}$ into an equisatisfiable 
set of \emph{initial clauses} $\mathcal{C}$.
It then tries to derive a contradiction $\bot$ from $\mathcal{C}$ (thus showing that $A_1,\ldots,A_l \models G$)
using a logical calculus, such as resolution or superposition \cite{DBLP:books/el/RV01/BachmairG01,DBLP:books/el/RV01/NieuwenhuisR01}.
The employed process of iteratively deriving (according to the inference rules of the calculus) new clauses, logical consequences of $\mathcal{C}$,
is referred to as \emph{saturation} and is typically implemented using some variant of 
a \emph{given-clause algorithm} \cite{DBLP:journals/jsc/RiazanovV03}: in each iteration, a single clause $C$ is $\emph{selected}$ and 
inferences are performed between $C$ and all previously selected clauses.
Deciding which clause to select next is known to be a key heuristical choice point,
hugely affecting the performance of an ATP \cite{DBLP:conf/cade/SchulzM16}.

The idea to improve clause selection by learning from past prover experience goes (to the best of our knowledge) back to Schulz \cite{Schulz:Diss-2000,DS1996b}
and has more recently been successfully employed by the ENIGMA system 
\cite{DBLP:conf/mkm/JakubuvU17,DBLP:conf/mkm/JakubuvU18,DBLP:conf/cade/ChvalovskyJ0U19,DBLP:conf/cade/JakubuvCOP0U20}
and variations \cite{DBLP:conf/lpar/LoosISK17,DBLP:journals/corr/abs-2006-11259,DBLP:journals/corr/abs-1911-02065}.
Experience is collected from successful prover runs, where each selected clause constitutes a training example and the example is marked as \emph{positive}
if the clause ended up in the discovered proof, and \emph{negative} otherwise.
A machine learning (ML) algorithm is then used to \emph{fit} this data and produce a \emph{model} $\mathcal{M}$
for \emph{classifying} clauses into positive and negative, accordingly.
A good learning algorithm produces a model $\mathcal{M}$ which accurately
classifies the training data but also \emph{generalizes} well to unseen examples;
ideally, of course, with a low computational cost of both (1) training and (2) evaluation.

When a model is prepared, we need to \emph{integrate} its advice back to the prover's 
clause selection process. An ATP typically organizes this process by maintaining 
a set of priority queues, each ordering the yet-to-be-processed clauses by a certain \emph{criterion},
and alternates---under a certain configurable \emph{ratio}---between selecting the best clause from each queue.
One way of integrating the learnt advice, adopted by ENIGMA, is to add another queue $Q_\mathcal{M}$
in which clauses are ordered such that those positively classified by $\mathcal{M}$ 
precede the negatively classified ones, and extend the mentioned ratio such
that $Q_\mathcal{M}$ is used for, e.g., half of the selections (while the remaining ones fall back to the original strategy).

\label{sec:the_new_in_old}

In this work, we rely instead on the \emph{layered clause selection} paradigm \cite{DBLP:conf/ijcai/Gleiss020,DBLP:conf/cade/Gleiss020,DBLP:conf/cade/Tammet19},
in which a clause selection mechanism inherited from an underlying strategy is applied 
separately to the set $A$ of clauses classified as \emph{positive} by $\mathcal{M}$ and to the set $B$ of \emph{all} 
yet-to-be-processed clauses (i.e., $A \subseteq B$).
A ``second level'' ratio then dictates how often will the prover relay to select from either of these two sets.
For example, with a \emph{second-level ratio} 2:1, the prover will select twice from $A$ 
(unless $A$ is currently empty and a fallback to $B$ happens)
before selecting once from $B$. An advantage of this approach is that the original, 
typically well-tuned, selection mechanism is still applied within both $A$ and $B$.\footnote{
We compared and empirically evaluated various modes of integrating the model advice
into the clause selection process in our previous work \cite{SudaCADE2021}. }






\section{Neural Classification of Clause Derivations} 

\label{sec:neural}

In our previous work \cite{SudaCADE2021}, we introduced a method for classifying clauses for clause selection
based on their derivation history, i.e., ignoring the logical content of the clauses. 
Each clause is characterised by the initial clauses from which it was derived, the inference rules
by which it was derived, and the exact way in which the rules were used to derive it.
The method relies on a recursive neural network (RvNN) as the machine learning architecture,
with the recursion running ``along'' the clause derivations, starting off from the initial clauses.

The previous work \cite{SudaCADE2021} focused on the \smtlib{} benchmark \cite{BarFT-SMTLIB} and on the aspect of theory reasoning
in \vampire{} implemented by adding theory axioms to formalise various theories of interest 
(arithmetic, arrays, data structures, \ldots). Together with the actual problem formulation, i.e., the user-supplied axioms,
the theory axioms become the initial clauses to start off the recursion for a RvNN.
Because these theory axioms are added by \vampire{} itself,
their use in derivations can be traced and meaningfully compared across problems from different sources
such as those comprising \smtlib{}. 

In this paper, we adapt the method to work with a different benchmark, 
namely, the Mizar40 \cite{DBLP:journals/jar/KaliszykU15a} problem set (see Sect.~\ref{sec:exper} for more details).
There is no explicit theory reasoning (in the sense of ``satisfiability modulo theories'') needed to solve problems in this benchmark.
On the other hand, many axioms appear in many problems across the Mizar40 benchmark and they are consistently named.
We rely on these consistently named axioms to seed the recursion here.

In this section, we first recall the general RvNN architecture for learning from clause derivations.
We then explain how this can be enhanced by incorporating information about the conjecture
(which is missing in \smtlib{}). We avoid repeating the technical details mentioned previously \cite{SudaCADE2021},
but include a subsection on our parallel training setup that we believe is of independent interest.

\subsection{A Recursive Neural Network Clause Derivation Classifier} \label{sec:rvnn_proper}

A recursive neural network (RvNN) is a network created by composing 
a finite set of neural building blocks \emph{recursively} over a structured input \cite{DBLP:conf/icnn/GollerK96}.

In our case, the structured input is a clause derivation: a directed acyclic (hyper-)graph (DAG)
with the initial clauses $C \in \mathcal{C}$ as leaves and the derived clauses as internal nodes, 
connected by (hyper-)edges labeled by the corresponding applied inference rules.
To enable the recursion, an RvNN represents each node $C$ by a real vector $v_C$ (of fixed dimension $n$)
called a (learnable) \emph{embedding}. During training, our network learns to embed the space of derivable clauses into $\mathbb{R}^n$
in a priori unknown, but hopefully reasonable way. 



We assume that each initial clause $C$ can be identified with an axiom $A_C$ from which it was obtained through clausification
(unless it comes from the conjecture) and that these axioms form a finite set $\mathcal{A}$, fixed for the domain of interest.
Now, the specific building blocks of our architecture are (mainly; see below) the following three (indexed families of) functions: 
\begin{itemize}
\item
	for every axiom $A_i \in \mathcal{A}$, a nullary \emph{init} function $I_i \in \mathbb{R}^n$ 
	which to an initial clause $C \in \mathcal{C}$ obtained through clausification from the axiom $A_i$ assigns its embedding $v_C := I_i,$
\item
	for every inference rule $r$, a \emph{deriv} function, $D_r : \mathbb{R}^n \times \cdots \times \mathbb{R}^n \to \mathbb{R}^n$ which
	to a conclusion clause $C_c$ derived by $r$ from premises $(C_1,\ldots,C_k)$ with embeddings $v_{C_1},\ldots,v_{C_k}$ 
	assignes the embedding $v_{C_c} := D_r(v_{C_1},\ldots,v_{C_k})$,
\item
	and, finally, a single \emph{eval} function $E : \mathbb{R}^n \to \mathbb{R}$ 
	which evaluates an embedding $v_C$ such that
	the corresponding clause $C$ is classified as \emph{positive} whenever $E(v_C) \geq 0$ and negative otherwise.
\end{itemize}
By recursively composing these functions, any derived clause $C$ can be 
assigned an embedding $v_C$ and evaluated to see whether the network recommends it as positive,
that should be preferred in the proof search, or negative, which will (according to the network) not likely contribute to a proof.
Notice that the amortised cost of evaluating a single clause by the network is low,
as it amounts to a \emph{constant number} of function compositions.



\subsection{Information Sources, the Conjecture, and SInE Levels}

\label{subsect:levels}

Let us first spend some time here to consider what kind of information about a clause
can the network take into account to perform its classification.

The assumption about a fixed axiom set enables meaningfully carrying between problems
observations about which axioms and their combinations
quickly lead to good lemmas and which are, on the other hand, 
 rarely useful.
We believe this is the main source of information for the network to classify well.
However, it may not be feasible to represent in the network all the axioms 
available in the benchmark. It is then possible to only \emph{reveal} a specific
subset to the network and represent the remaining ones using 
a single special embedding $I_\mathit{unknown}$.

Another, less obvious, source of information are the inference rules. Since there are distinct 
\emph{deriv} functions $D_r$ for every rule $r$, the network can also 
take into account that different inference rules give rise to conclusions of 
different degrees of usefulness. In Sect.~\ref{subsect:breakdown}, we dedicate an experiment to establishing how much this aspect 
of the architecture helps clause classification.

Finally, we always ``tell the network'' what the current conjecture $G$ is
by marking the conjecture clauses using a special initial embedding $I_\mathit{goal}$.\footnote{
By special, we mean ``in principle distinct''. Since all the embeddings are learnable,
the network itself ``decides'' during training how exactly to distinguish $I_\mathit{goal}$
and all the other axioms embeddings $I_i$ (and also the ``generic'' $I_\mathit{unknown}$).}
Focusing search on the conjecture is a well-known theorem proving heuristic 
and we give the network the opportunity to establish how strongly should 
this heuristic be taken into account. 

We actually implemented a stronger version of the conjecture-focus idea
by precomputing (and incorporating into the network) for every initial clause its SInE level \cite{DBLP:conf/ijcai/Gleiss020,Vampire2019:Aiming_for_Goal_with,DBLP:conf/cade/HoderV11}.
A SInE level is a heuristical distance of a formula from the conjecture along a relation defined 
by sharing signature symbols \cite{Vampire2019:Aiming_for_Goal_with}. Roughly, the SInE levels are
computed as a byproduct of the iterative SInE premise selection algorithm \cite{DBLP:conf/cade/HoderV11},
where we assign the level $l$ to axiom $A$ if the SInE algorithm first considers adding axiom $A$
among the premises in its $l$-th iteration. Thus, the conjecture itself is assigned level 0,
and a typical configuration of the algorithm on a typical formula (as witnessed by our experiments)
assigns levels between 1 to around 10 to the given axioms. 

To incorporate SInE levels into our network, we pass the embedding $I \in \mathbb{R}^n$ produced by an init function 
through an additional (learnable) \emph{SInE embedder} function $S: \mathbb{R}^n \times \mathbb{R} \to \mathbb{R}^n$.
Thus, an initial clause $C \in \mathcal{C}$ obtained through clausification from the axiom $A_i$ and with a SInE level $l$
receives an embedding $S(I_i,l) \in \mathbb{R}^n$. Also the effect of enabling or disabling this extension is demonstrated
in the experiments in Sect.~\ref{subsect:breakdown}.

\subsection{Training the Network} \label{subsect:training_theory}

Our RvNN is parametrized by a tuple of \emph{learnable parameters} $\Theta = (\theta^I,\theta^D,\theta^E,\theta^S)$ 
which determine the corresponding init, deriv, eval, and SInE embedder functions (please consult our previous
work \cite{SudaCADE2021}, Sect.~4.2, for additional details). 
To train the network means  to find suitable values for these parameters such that it successfully classifies
positive and negative clauses from the training data and ideally also generalises to unseen future cases.

We follow a standard methodology for training our networks. In particular, we use the gradient descent (GD) optimization algorithm 
minimising a binary cross-entropy \emph{loss} 
\cite{DBLP:books/daglib/0040158}.
Every clause in a derivation DAG selected by the saturation algorithm
constitutes a contribution to the loss, with the clauses that participated in the found proof receiving
the target label $1.0$ (positive example) and the remaining ones the label $0.0$ (negative example).
We \emph{weight} these contributions such that each derivation DAG (corresponding to a prover run on a single problem)
receives equal weight, and, moreover, within each DAG we scale the importance of positive and negative 
examples such that these two categories contribute evenly.

We split the available successful derivations into \SI{90}{\percent} \emph{training} set and \SI{10}{\percent} \emph{validation} set,
and only train on the first set using the second to observe generalisation to unseen examples.
As the GD algorithm progresses, iterating over the training data in rounds called \emph{epochs},
typically, the loss on the training examples steadily decreases while the loss on the validation set
at some point stops improving or even starts getting worse. In our experiments, we always pick the model
with the smallest validation loss for evaluation with the prover, as these models were shown
to lead to the best performance in our previous work \cite{SudaCADE2021}.

\subsection{Implementation and the Parallel Training Setup}

\label{subsect:train_in_par}

We implemented an infrastructure for training an RvNN clause derivation classifier in Python, using the PyTorch (version 1.7) 
library \cite{NEURIPS2019_9015} and its {Torch\-Script} extension for later interfacing the trained model from C++.\footnote{
The implementation is available as a public repo at \url{https://git.io/JOh6S}.}

PyTorch is built around the concept of dynamic computational graphs 
for the calculation of gradient values required by the GD algorithm.
This is an extremely flexible approach in which the computational graph is automatically constructed
while executing code that looks like simply performing the vector operations pertaining
to evaluating the network's concrete instance on a concrete set of training examples (corresponding, in our case, to a concrete clause derivation).  
A downside of this approach is that the computational graph cannot be stored and reused
when the same example is to be evaluated and used for training in the next epoch.
As a consequence, most of the time of training an RvNN like ours is spent on constructing 
computational graphs over and over again.

\paragraph{Batching:} One general way of speeding up the training of a neural network amounts to grouping training examples 
into reasonably sized sets called \emph{batches} and processing them in parallel---in the sense of single instruction multiple data (SIMD)---typically on a specialized hardware such as a GPU. However, this is most easily done only when the training examples have the same
shape and can be easily aligned, such as, e.g., with images, and is not immediately available with RvNNs.\footnote{
There exist non-trivial preprocessing techniques for achieving graph batching \cite{DBLP:conf/iclr/LooksHHN17}. }

Because each clause derivation that we want to process in training is in general of a unique shape,
we do not attempt to align multiple derivations to benefit from SIMD processing. Instead, we create 
batches by merging multiple derivations to simply create DAGs of comparable size (some derivations
are relatively small, while the largest we encountered were of the order of hundred thousand nodes).
%
%
By merging we mean: (1) putting several derivations next to each other, and (2) identifying and collapsing
nodes that are indistinguishable from the point of view of the computation our RvNN performs.\footnote{
The latter is already relevant within a single derivation (c.f. \cite{SudaCADE2021}, Sect.~4.4).}
This means that a batch contains at most one node for all initial clauses
corresponding to the conjecture or at most one node for all clauses derived from
two such initial clauses in a single step by resolution, etc. When collapsing nodes from
different derivations, we make sure to compute the correct target labels and their weights to preserve the semantics
the network had before the merge.\footnote{
E.g., a node can be designated a positive example (label $1.0$, weight $w_1$) in one derivation and a negative one (label $0.0$, weight $w_2$) in another. 
The corresponding collapsed node receives the label $w_1/(w_1+w_2)$ and weight $(w_1+w_2)$.}

\paragraph{A multi-process training architecture:} 
To utilise parallelism and speed up the training in our case (i.e., with similarly sized but internally heterogenous batches),
we implemented a master-worker multiprocess architecture to be run on a computer with multiple CPUs (or CPU cores).

The idea is that a master process maintains a single official version of the network (in terms of the learnable parameters $\Theta$)
and dispatches training tasks to a set of worker processes. A training task is a pair $(\Theta_t,B)$ where $\Theta_t$
is the current version of the network at time $t$ (the moment when the task is issued by the master process) and $B$ is a selected batch.
A worker process constructs the computation graph corresponding to $B$, performs a back-propagation step,
and sends the obtained gradient $\nabla \Theta_t(B)$ back to the master. The master dispatches tasks and receives gradients from finished workers
using two synchronization queues. The master updates the official network after receiving a gradient from a worker via
\[\Theta_{T+1} \leftarrow \Theta_T - \alpha \nabla \Theta_{t}(B), \]
where $\alpha$ is the learning rate. 

A curious aspect of our architecture is that $\Theta_T$, the network the master updates at moment $T$
using $\nabla \Theta_t(B)$, is typically a later version than $\Theta_t$ from which the corresponding task has been derived.
In other words, there is a certain drift between the version of the network an update has been computed for 
and the version the update is eventually applied to. This drift arises because the master issues a task as soon
as there is a free worker and receives an update as soon as there is a finished worker. In a sense,
such drift is necessary if we want to keep the workers busy and capitalise on parallelisation at all.

Surprisingly, the drift seems to have a beneficial effect on learning in the sense that 
it helps to prevent overfitting. Indeed, we were able to train models with slightly smaller validation 
loss using parallelisation than without it.\footnote{
This effect has already been observed by researches in a related context \cite{NIPS2011_218a0aef}.}

\section{Experiments} \label{sec:exper}


We implemented clause selection guidance (Sect.~\ref{sec:enigma_style}) by a recursive neural network classifier for clause derivations (Sect.~\ref{sec:neural})
in the automatic theorem prover \vampire{} (version 4.5.1).\footnote{
Supplementary materials for the experiments can be found at \url{https://git.io/JOY71}.}
In our previous work \cite{SudaCADE2021}, we used the \smtlib{} benchmark \cite{BarFT-SMTLIB}
and empirically compared several modes of integration of the learned advice into the prover
as well as several supporting techniques. In this paper, we set out with the best configuration
identified therein\footnote{This means layered clause selection with second-level ratio 2:1 (as explained in Sect.~\ref{sec:enigma_style})
and lazy model evaluation and abstraction caching (see \cite{SudaCADE2021}).}
and focus on evaluating various aspects of the neural architecture itself
and do that using the problems from the Mizar mathematical library.


Following Jakub\r{u}v and Urban \cite{DBLP:conf/itp/JakubuvU19}, we use the Mizar40 \cite{DBLP:journals/jar/KaliszykU15a} benchmark
consisting of \num{57880} problems from the MPTP  \cite{DBLP:journals/jar/Urban06} and, in particular, the small (\emph{bushy}, re-proving) version.
This version emulates the scenario where some form of premise selection 
has already occurred and allows us to directly focus on evaluating internal guidance in an ATP.
To allow for a direct comparison with Jakub\r{u}v and Urban's remarkable results \cite{DBLP:conf/itp/JakubuvU19},
we adopt the base time limit of \SI{10}{\second} per problem
and use comparable hardware for the evaluation.\footnote{A server with Intel(R) Xeon(R) Gold 6140 CPUs @ \SI{2.3}{\giga\hertz} with \SI{500}{\giga\byte} RAM.
}


This section has several parts. First, we explain the details concerning the initial run from which 
the training derivations got collected, describe the various ways of setting up the training procedure we experimented with,
and evaluate the performance of the obtained models (Sects~\ref{subsect:prepare}--\ref{subsect:primary_eval}).
We then set out to establish how the individual aspects of our architecture (as discussed in Sect.~\ref{subsect:levels})
contribute to the overall performance (Sect.~\ref{subsect:breakdown}).
Finally, we follow Jakub\r{u}v and Urban \cite{DBLP:conf/itp/JakubuvU19} in training better and better models
using the growing set of solved problems for more training and compare with their results (Sect.~\ref{subsect:looping}).

%
%
%
%
%

\subsection{Data Preparation}

\label{subsect:prepare}

We first identified a \vampire{} strategy (from among \vampire{}'s standard CASC schedule) which performed well on the Mizar40 benchmark.
We denote the strategy here as $\mathcal{V}$ and use it as a baseline.
This strategy solved a total of \num{20197} problems under the base \SI{10}{\second} time limit.

The corresponding successful derivations amount to roughly \SI{800}{\mega\byte} of disk space when zipped.
There are \num{43080} named Mizar axioms occurring in them (and in each also some conjecture clauses),
and \num{12} inference rules including resolution, factoring, superposition, forward and backward demodulation, subsumption resolution,
unit resulting resolution and AVATAR (which represents the connection between a clause getting split and its components) \cite{DBLP:conf/cav/Voronkov14,DBLP:conf/cade/RegerSV15}. 

As the total number of axioms seemed too large, we only took a subset of size $m$ of the most often occurring ones 
to be represented by distinct labels for the network to distinguish and replaced the remaining ones in the dataset by the single generic label $\mathit{unknown}$ (c.f. $I_\mathit{unknown}$ in Sect.~\ref{subsect:levels}).
To evaluate the effect of this hyper-parameter on performance, we initially
experimented with three values of $m$, namely, $500,1000$, and $2000$.

To finalize the preparation of the training dataset(s), we constructed merged batches of approximately \num{20} thousand nodes each.
71 derivations were larger than this threshold (the largest had \num{242023} merged nodes), but a typical batch merged 5--20 derivations.
In the particular case of $m=1000$, which we consider the default, we constructed \num{2167} batches in total.
We then randomly split these, as mentioned, into \SI{90}{\percent} of training and \SI{10}{\percent} of validation examples.

\subsection{Training} 

We trained our models using the parallel setup with 20 cores, for up to 100 epochs,
in the end choosing the model with the best validation loss as the result. 
Similarly to our previous work \cite{SudaCADE2021}, we used a variable learning rate.\footnote{
The learning rate was set to grow linearly from 0 to a maximum value $\alpha_m = \SI{2.0e-4}{}$ in epoch 40: $\alpha(t) = t\cdot \alpha_m/40$
for $t \in (0,40]$; and then to decrease from that value as the reciprocal function of time: 
$\alpha(t) = 40\cdot\alpha_m/t$ for $t\in (40,100).$}

In total, we ran five independent training attempts. In addition to the number of revealed axioms $m$,
we also set out to evaluate how the behaviour of the network
changes with the size of the embedding $n$. For the default number of revealed axioms $m=1000$,
we tried the embedding sizes $n = 64,128,$ and $256$.

\begin{table}[t]
\caption{Training statistics of models in the first experiment.}
\label{tab:loop0-model-statistics}
\centering
\setlength{\tabcolsep}{5pt}
\begin{tabular}{l||r|rrr|r}
model shorthand & $\mathcal{H}^{n128}$ & $\mathcal{M}^{n64}$ & $\mathcal{M}^{n128}$ & $\mathcal{M}^{n256}$ & $\mathcal{D}^{n128}$ \\
\hline
revealed axioms $m$ & \phantom{0.}500 & \phantom{.}1000 & \phantom{.}1000 & \phantom{.}1000 & \phantom{.}2000 \\
embedding size $n$ & \phantom{0.}128 & \phantom{00.}64 & \phantom{0.}128 & \phantom{0.}256 & \phantom{0.}128 \\
\hline
wall training time per epoch (\si{\minute}) & \phantom{0.0}42 & \phantom{0.0}32 & \phantom{0.0}48 & \phantom{0.0}74 & \phantom{0.0}58 \\ 
model size (\si{\mega\byte}) & \phantom{00}4.6 & \phantom{00}1.6 & \phantom{00}5.0 & \phantom{0}17.9  &  \phantom{00}5.8 \\
\hline
overall best epoch & \phantom{0.0}69 & \phantom{0.0}52 & \phantom{0.0}60 & \phantom{0.0}60 & \phantom{0.0}46 \\
validation loss    & 0.475 &
                     0.455 & 
                     0.455 & 
                     0.452 &
                     0.467 \\
true positive rate & 0.947 & 0.952 & 0.954 & 0.949 & 0.948 \\ 
true negative rate & 0.872 & 0.870 & 0.868 & 0.874 & 0.858 \\ 
\end{tabular}
\end{table}
The training statistics of the corresponding five models are summarized in Table~\ref{tab:loop0-model-statistics}.
For each of the tried combination of $m$ and $n$, the table starts by giving a shorthand to the best obtained model for later reference.

The next block documents the speed of training and the size of the obtained models.
We can see that the model sizes are dictated mainly by the embedding size $n$ and not so much by the number of revealed axioms $m$.
(Roughly, $\Theta(n^2)$ of space is needed for storing the matrices representing the deriv and eval functions,
while $\Theta(n \cdot m)$ space is required for storing the axiom embeddings.)
We note that the sizes are comparable to those of the gradient boosted trees used by Jakub\r{u}v and Urban \cite{DBLP:conf/itp/JakubuvU19} 
(\SI{5.0}{\mega\byte} for a tree of size 9 in their main experiment).
Concerning the training times, the \SI{48}{\minute} per epoch recorded for $\mathcal{M}^{n128}$ corresponds in 100 epochs to approximately 3 days
of 20 core computation and almost 70 single-core days. Jakub\r{u}v and Urban \cite{DBLP:conf/itp/JakubuvU19} trained a similarly sized model
in under 5 single-core days, which indicates that training neural networks is much more computation intensive.

Finally, Table~\ref{tab:loop0-model-statistics} also reports for each training process the epoch in which 
the validation loss was in the end the lowest, the achieved validation loss
and the (weighted) true positive and negative rates (on the validation examples).\footnote{
Please note that the batches of training and validation examples for different numbers of revealed axioms
were constructed and split independently, 
so meaningful comparisons are mainly possible between the values of the middle column (for $m = 1000$).}
The true positive rate (TPR) is the fraction of positive examples that the network identifies as such.
The true negative rate (TNR) is defined analogously. In our case, we use the same weighting formula
as for computing the contributions of each example to the loss (recall Sect.~\ref{subsect:training_theory}).
It is interesting to observe that on the Mizar benchmark here the training process automatically produces models 
biased towards better TPR (c.f. \cite{SudaCADE2021}, Sect.~5.5), 
while the weighting actually strives for an equal focus on the positive and the negative examples.

\subsection{Evaluation with the Prover}

\label{subsect:primary_eval}

\begin{table}[t]
\caption{Performance statistics of the base strategy $\mathcal{V}$ and
five strategies enhancing $\mathcal{V}$ with a clause selection guidance by the respective neural models from Table~\ref{tab:loop0-model-statistics}.}
\label{tab:loop0-1}
%
\centering
\setlength{\tabcolsep}{5pt}
\begin{tabular}{r||c||c|ccc|c}
strategy & 
	$\mathcal{V}$ & 
	$\mathcal{H}^{n128}$ & 
	$\mathcal{M}^{n64}$ & 
	$\mathcal{M}^{n128}$ & 
	$\mathcal{M}^{n256}$ &
	$\mathcal{D}^{n128}$ \\
\hline 
solved &
	 \num{20197} &
	 \num{24581} & 
	 \num{25484} & 
	 \num{25805} & 
	 \num{25287} &
{\bf \num{26014}} \\
$\mathcal{V}\%$ &
	 +\SI{0}{\percent} &
	 +\SI{21.7}{\percent} &
	 +\SI{26.1}{\percent} &
	 +\SI{27.7}{\percent} &
	 +\SI{25.2}{\percent} &	 
{\bf +\SI{28.8}{\percent}} \\
$\mathcal{V}+$ & 
	+0 & 
	+5022 &
	+5879 &
	+6129 &	
	+5707 &		
{\bf +6277} \\	
$\mathcal{V}-$ & 
	$-0$ &
	\phantom{0}$-638$ &
	\phantom{0}$-592$ &
	\phantom{0}$-521$ &
	\phantom{0}$-617$ &	
    {\bf $\phantom{0}\num{-460}$}\\
\hline
model eval. time & 
	\SI{0}{\percent} &
	\SI{37.1}{\percent} &
	{\bf \SI{32.9}{\percent}} &
	\SI{37.7}{\percent} &	
	\SI{48.6}{\percent} &	
	\SI{36.7}{\percent} \\
%
\end{tabular}
\end{table}

Next, we reran \vampire{}'s strategy $\mathcal{V}$, now equipped with the obtained models for guidance,
again using the time limit of \SI{10}{\second}. 
The results are shown in Table~\ref{tab:loop0-1}.

We can see that the highest number of problems is solved with the help of $\mathcal{D}^{n128}$,
the model with the intermediate embedding size $n=128$ but with the largest tried number of 
revealed axioms $m=2000$. The strategy equipped with $\mathcal{D}^{n128}$ solves \num{26014} problems,
which is \SI{28.8}{\percent} of the baseline $\mathcal{V}$.

In addition to the solved counts and the percentages, Table~\ref{tab:loop0-1} also shows the number of gained ($\mathcal{V}+$)
and lost ($\mathcal{V}-$) problems with respect to the base strategy $\mathcal{V}$.
Note that the problems from $\mathcal{V}+$ were not present in the training set, 
so solving those is a sign of successful generalization. On the other hand,
the non-negligible number of no-longer-solved problems under $\mathcal{V}-$   
reminds us of the overhead connected with interfacing the network.

The last row of the table elaborates on this, presenting the average time spent by the strategies 
on evaluating their respective models.
The numbers indicate that the evaluation time is mainly determined by the embedding size $n$,
as the models with $n=128$ all spend approximately \SI{37}{\percent} on evaluating,
while notable differences appear with $n$ getting varied.

It is now interesting to compare the evaluation time (i.e., how fast the advice is) and
the validation loss from Table~\ref{tab:loop0-model-statistics} (i.e., how good the advice is)
with the observed ATP performance. It appears that $\mathcal{M}^{n256}$ is too slow 
to capitalize on its superior advice quality over $\mathcal{M}^{n64}$ and $\mathcal{M}^{n128}$.
However, there must be limits to how indicative the validation loss is for the final performance,
because this metric does not help distinguish between $\mathcal{M}^{n64}$ and $\mathcal{M}^{n128}$
and yet the slower to evaluate $\mathcal{M}^{n128}$ eventually helps \vampire{} solve substantially more problems.

%

\subsection{Information Source Performance Breakdown}

\label{subsect:breakdown}

So far, we observed how the performance of the guided prover changes 
when we vary the two numerical parameters of our architecture,
namely, the size of the embedding $n$ and the number of revealed axioms $m$.
Here we want to shed more light on how the performance arises from the contributions
of our architecture's main information sources: the ability to distinguish the input axioms
at all (i.e., the information channelled through the init functions),
the ability to distinguish individual inference rules (corresponding to the deriv functions),
and the ability to track relatedness to the conjecture via the SInE levels.
We do this by disabling 
these sources in turn and rerunning the prover.

\paragraph{No distinguished input axioms.}
With each information source, we have in principle two options. One option is to train 
a new network from scratch, but on a dataset which does not contain the extra information 
corresponding to the disabled source (e.g., with $m=0$ revealed axioms). The other option is
to use an already trained model (we will use $\mathcal{M}^{n128}$ for this),
but to withhold the extra information while evaluating the model in the prover.
For this, we need to provide a default value for ``masking out'' the extra information.
(When disabling input axioms, we simply use $I_\mathit{unknown}$ as the default
to embed any input clause except the conjecture ones.) Note that the two options 
are not equivalent and, intuitively, the first one should not perform worse than the second.\footnote{
The first option is like being born blind, learning during life how to live without the missing sense,
the second option is like losing a sense ``just before the final exam''.}

\begin{table}[t]
\caption{Performance decrease when 
no axiom information is a available ($\mathcal{A}^0$ and $\mathcal{M}_\mathit{noAx}$)
and 
when inference rules are not distinguished ($\mathcal{R}_\mathit{defR}$). 
All models used $n=128$ and $\mathcal{M}$ stands for $\mathcal{M}^{n128}$ from Tables~\ref{tab:loop0-model-statistics} and \ref{tab:loop0-1}.
Further details in the main text.}
\label{tab:building-blocks-I}
\centering
\setlength{\tabcolsep}{3pt}
\begin{tabular}{r||c|cc|cc}
strategy & 
  $\mathcal{M}$ &  
  $\mathcal{A}^0$ & 
  $\mathcal{M}_\mathit{noAx}$ &
  $\mathcal{R}$ &
  $\mathcal{R}_\mathit{defR}$ \\
\hline
solved 
	& \num{25805} & 
	\num{21400} &
	\num{21011} &
	\num{25686} &
	\num{24544} \\
$\mathcal{M}\%$ & 
	 \phantom{0}+\SI{0.0}{\percent} &
	 $-$\SI{17.0}{\percent} &
	 $-$\SI{18.5}{\percent} &
	 \phantom{0}$-$\SI{0.4}{\percent} &
	 \phantom{0}$-$\SI{4.8}{\percent} \\
\hline
$\mathcal{V}\%$ & 
    +\SI{27.7}{\percent} &
	\phantom{0}+\SI{5.9}{\percent} &
	\phantom{0}+\SI{4.0}{\percent} & 
	+\SI{27.1}{\percent} &
	+\SI{21.5}{\percent} \\	
\end{tabular}
\end{table}

%

We can observe the effect of disabling access to the input axioms in Table~\ref{tab:building-blocks-I}.
Model $\mathcal{A}^0$ represents the just described option one, where the axiom information was already witheld during training.
The column $\mathcal{M}_\mathit{noAx}$, on the other hand, used the original model $\mathcal{M}^{n128}$ (here dubbed simply $\mathcal{M}$),
but during evaluation in the prover all axioms were deliberately presented as $\mathit{unknown}$.

Most important to notice is that both options perform much worse than the original model $\mathcal{M}$,
which shows that the ability to distinguish the input axioms is crucial for the good performance of our architecture.
Nevertheless, when compared to the baseline strategy $\mathcal{V}$, the guided prover still solves around \SI{5}{\percent}
more problems. This means the guidance
is still reasonably good (given that almost \SI{40}{\percent} of the proving time is spent evaluating the network).
Finally, $\mathcal{A}^0$ performs slightly better than $\mathcal{M}_\mathit{noAx}$, which conforms with our intuition.

\paragraph{No distinguished derivation rules.}

%

There seems to be no obvious way to pick a default inference rule for masking out the functionality of this information source.
The situation is further complicated by the fact that we need at least two defaults based on the inference rule arity.\footnote{
Out architecture separately models arity one rules, binary rules, and rules with arity of 3 and more for which a binary building
block is iteratively composed with itself.}
To prepare such defaults,\footnote{These could also be used whenever a trained model is combined with a strategy not used to produce
the training data, 
possibly invoking rules not present in training.}
we came up with a modification of the training regime that we call \emph{swapout}.\footnote{
In honor of \emph{dropout} \cite{DBLP:journals/jmlr/SrivastavaHKSS14}, a well-know regularization technique that inspired this. }

Training with swapout means there is a nonzero probability $p$ (we used $p=0.1$ in the experiment)
that an application of a particular inference rule $r$ in a derivation---i.e.~applying the deriv function $D_r$ to produce 
the next clause embedding---will instead use a generic function $D_{\mathit{arity}(r)}$ shared by all rules of the  
same arity as $r$. This is analogous to using $I_\mathit{unknown}$ for axioms not important enough 
to deserve their own init function, but decided probabilistically.

The right part of Table~\ref{tab:building-blocks-I} presents the performance of a model obtained using swapout.
First, under $\mathcal{R}$, the additionally trained generic deriv functions were ignored. 
This means that $\mathcal{R}$ uses the same full set of information sources as $\mathcal{M}$.
We should remark that training with swapout took longer to reach the minimal validation loss and the final loss
was lower than that of $\mathcal{M}$ (0.454 in epoch 95). Nevertheless, $\mathcal{R}$ performs slightly worse than $\mathcal{M}$.


Under $\mathcal{R}_\mathit{defR}$, we see the performance of $\mathcal{R}$
where the trained generic deriv functions are exclusively used to replace (based on arity) the specific ones. 
The performance drops by approximately \SI{5}{\percent}
compared to $\mathcal{M}$, which shows that there is value in the architecture being able to distinguish the derivation rules.

\paragraph{No SInE levels.} 

\begin{table}[t]
\caption{The effect of training without SInE levels information ($\mathcal{S}^0$)
and of imposing various fixed SInE levels on $\mathcal{M} = \mathcal{M}^{n128}$.}
\label{tab:building-blocks-II}
\centering
\setlength{\tabcolsep}{2pt}
\begin{tabular}{r||c|c|cccccc}
strategy & 
  $\mathcal{M}$ &
  $\mathcal{S}^0$ &
  $\mathcal{M}_{l=0}$ &
  $\mathcal{M}_{l=1}$ &
  $\mathcal{M}_{l=2}$ &
  $\mathcal{M}_{l=3}$ &  
  $\mathcal{M}_{l=4}$ & 
  $\mathcal{M}_{l=5}$ \\
\hline
solved & 
	\num{25805} &
	\num{25440} &
	\num{25724} &
	\num{25823} &
	\num{25882} &
	{\bf \num{25884}} &
	\num{25866} &
	\num{25802} \\
$\mathcal{M}\%$ & 
	 \phantom{0}+\SI{0.0}{\percent} &
	 \phantom{0}$-$\SI{1.4}{\percent} &
	 \phantom{0}$-$\SI{0.3}{\percent} &
	 \phantom{0}+\SI{0.0}{\percent} &
	 \phantom{0}+\SI{0.2}{\percent} &
	{\bf \phantom{0}+\SI{0.3}{\percent}} &
	 \phantom{0}+\SI{0.2}{\percent} &
	 \phantom{0}$-$\SI{0.0}{\percent} \\
\hline
$\mathcal{V}+$ &  
	{\bf+6129 }&
	+5783 &
	+5878 &
	+6002 &
	+6092 &
	+6101 &
	+6114 & 
	+6108 \\
$\mathcal{V}-$ & 
	\phantom{0}$-521$ &
	\phantom{0}$-540$ &	
	{\bf \phantom{0}$\num{-351}$} &
	\phantom{0}$-376$ &
	\phantom{0}$-407$ &
	\phantom{0}$-414$ &
	\phantom{0}$-445$ &
	\phantom{0}$-503$ \\
\end{tabular}
\end{table}

%

Let us finally move to the information source provided by the SInE levels 
and a corresponding experiment documented in Table~\ref{tab:building-blocks-II}.
In that table, $\mathcal{S}^0$ is a model trained without access to this source, 
while the remaining columns represent $\mathcal{M}$ with increasingly large values 
of the SInE level $l$ uniformly hardwired for evaluation in the prover. 

Confusingly, all $\mathcal{M}$-derived models fare better than $\mathcal{S}^0$
and some of them are even better than $\mathcal{M}$ itself.
We currently do not have a good general explanation for this phenomenon,
although an analogy with the success of ``positive bias'' observed in our previous work on \smtlib{} can be drawn (c.f.~\cite{SudaCADE2021}, Sect.~5.5).
Hardwiring a low SInE level $l$ means the network will consider many clauses to be more related to the conjecture
than they actually are, which will likely lead to more clauses classified as positive.
Then the general intuition would be that it is more important for performance 
not to dismiss a clause needed for the proof than to dismiss clauses that will not be needed.

It is worth pointing out that $\mathcal{M}_{l=0}$ is the most ``careful'' configuration of these,
scoring the lowest in terms of $\mathcal{V}-$, the number of problems 
lost with respect to the baseline strategy $\mathcal{V}$. 
Additionally, $\mathcal{M}$ still scores the highest on $\mathcal{V}+$, the number of newly solved problems,
not present in the training data, although $\mathcal{M}_{l=4}$ comes quite close.
More analysis seems to be needed to fully understand the effect of the SInE levels on the architecture's performance.


\subsection{Looping to Get Even Better}

\label{subsect:looping}

When evaluating a strategy guided by a model leads to solving previously unsolved problems, 
the larger set of proofs may be used for training a potentially 
even better model to help solve even more problems.
Jakub\r{u}v and Urban \cite{DBLP:conf/itp/JakubuvU19} call this method \emph{looping}
and successfully apply it on Mizar for several iterations.

Here we report on applying looping to our neural architecture. We follow our previous work 
and adhere to the following two rules when using the method: First, we use exactly one successful derivation
to train on for every previously solved problem. Second, if a derivation was obtained with the help
of previously trained guidance, we augment the derivation with the unsuccessful run of plain $\mathcal{V}$ on that problem.
The first rule ensures the dataset does not grow too large too quickly.
The second rule helps to create a sufficient pool of negative examples, ``typical bad decisions'', that might 
otherwise not be present in a derivation obtained with some form of guidance already in place
(c.f.~\cite{SudaCADE2021}, Sect.~5.6).

\begin{table}[t]
\caption{Summary of the looping procedure. \emph{Collected} stands for the number of derivations available for training.
\emph{Performance} refers to the best strategy of the loop (in \SI{10}{\second}). }
\label{tab:looping}
\centering
\setlength{\tabcolsep}{3pt}
\begin{tabular}{c|cc|cccc}
             & \multicolumn{2}{c|}{training} & \multicolumn{3}{c}{evaluation} \\
loop index & collected & $m$ & performance & $\mathcal{V}\%$ & \%collected   \\
\hline 
0 &  ---  & ---                 & \num{20197} 
											& \phantom{0}+\SI{0.0}{\percent} & --- \\
1 & \num{20197} & 500/1000/2000 & \num{26014} 
											& +\SI{28.8}{\percent} & \SI{128.8}{\percent} \\
2 & \num{29065} & 3000          & \num{27348} & +\SI{35.4}{\percent} & \phantom{0}\SI{94.0}{\percent} \\
3 & \num{32020} & 5000          & \num{28947} & +\SI{43.3}{\percent} & \phantom{0}\SI{90.4}{\percent} \\
\end{tabular}
\end{table}






The results of looping are summarized in Table~\ref{tab:looping}. We can already recognize the values 
of the first two rows: ``Loop 0'' means the run of the baseline strategy $\mathcal{V}$.
Then, in loop 1, the obtained \num{20197} successful derivations become available for training (some actually get used
for training, others for validation), and, as we know from Table~\ref{tab:loop0-1}, 
the best model of this first round of training was $\mathcal{D}^{n128}$,
solving \num{26014} Mizar40 problems under the \SI{10}{\second} time limit.

For the next loop---observing the two rules mentioned above---we collected a total of \num{29065} successful 
derivations
and trained the next model using an increased number of revealed axioms $m = 3000$. To create additional variability in the runs and thus to increase 
the chances of collecting even more derivations for the next loop, we varied the modes of interfacing a model,
studied in more detail in our previous work \cite{SudaCADE2021}. The best configuration of loop 2
solved \num{27348} problems and the union of solved problems 
grew to \num{32020}. Finally, training using the corresponding successful derivations in loop 3,
we were able to produce a model $\mathcal{B}$ with $n=128$, $m=5000$ that can guide \vampire{}\footnote{
Using again the here prevalent layered clause selection with second-level ratio 2:1.}
to solve \num{28947} problems and thus improves over the baseline $\mathcal{V}$ by more than \SI{43}{\percent}.

As can be seen from Table~\ref{tab:looping}, while the best strategy's performance improves with every loop,
there is clearly an effect of diminishing returns at play. In particular, after loop 1 
the best strategy is no longer able to solve more problems than was the number of solutions used for training the corresponding model
and in loop 3 their percentage comparison 
(i.e., \%collected) only reaches \SI{90}{\percent}.
Another observation is that our initial estimate $m=1000$ for a reasonable number of revealed axioms
was too low. The additional capacity is paying off even for $\mathcal{B}$, which with its $m=5000$ 
reaches a size of \SI{8.8}{\mega\byte} (c.f.~Table~\ref{tab:loop0-model-statistics}) and \SI{40.1}{\percent} running time spent on model evaluation (c.f.~Table~\ref{tab:loop0-1}).

Let us conclude here by a comparison with the results of Jakub\r{u}v and Urban \cite{DBLP:conf/itp/JakubuvU19}.
They start off with a strategy of E~\cite{SCV:CADE-2019} solving \num{14933} Mizar40 problems under a \SI{10}{\second} time limit
and their best loop 4 model guides ENGIMA to solve \num{25397} problems (i.e., +\SI{70}{\percent}) under that time limit.
The authors kindly provided us with the precise set of problems solved by their runs.
Their runs cover \num{27425} problems. Our collection, that could be used for training in our next loop,
counts \num{32531} solved problems.
Our architecture solved \num{6356} problems that ENGIMA could not (and did not solve \num{1250} problems that ENIGMA could).


\section{Conclusion}

There is a new neural architecture for guiding clause selection
in saturation-based ATPs based solely on clause derivation history \cite{SudaCADE2021}.
We adapted this architecture to work in the context of a large library 
of formalized mathematics, in particular the Mizar mathematical library (MML) \cite{DBLP:journals/jfrea/GrabowskiKN10},
and conducted a series of experiments on the Mizar40 export of the library \cite{DBLP:journals/jar/KaliszykU15a}
with the new architecture interfaced from the ATP \vampire{}.
We established how the performance of the obtained system depends on  
parameters of the network and on its architectural building blocks.
We also compared its performance to that of ENIGMA 
and saw our architecture further improve on ENIGMA's remarkable results 
\cite{DBLP:conf/itp/JakubuvU19}.




It is perhaps surprising that so much can be gained by simply paying attention to the clause's pedigree
while ignoring what it says as a logical formula.
In future work, we would like to have a closer look at the trained models (and thus, implicitly, at the successful derivations)
and employ the techniques of explainable AI to get a better understanding of the architecture's success.
We hope to distill new general purpose theorem proving heuristics 
or, at least, contribute to knowledge transfer from Mizar to other libraries of formalized mathematics.







\section*{Acknowledgement}
This work was supported by the Czech Science Foundation project 20-06390Y
and the project RICAIP no. 857306 under the EU-H2020 programme.

\bibliographystyle{splncs04}
\bibliography{main}

\begin{thebibliography}{10}
\providecommand{\url}[1]{\texttt{#1}}
\providecommand{\urlprefix}{URL }
\providecommand{\doi}[1]{https://doi.org/#1}

\bibitem{DBLP:journals/jar/AlamaHKTU14}
Alama, J., Heskes, T., K{\"{u}}hlwein, D., Tsivtsivadze, E., Urban, J.: Premise
  selection for mathematics by corpus analysis and kernel methods. J. Autom.
  Reason.  \textbf{52}(2),  191--213 (2014). \doi{10.1007/s10817-013-9286-5}

\bibitem{DBLP:journals/corr/AlemiCISU16}
Alemi, A.A., Chollet, F., Irving, G., Szegedy, C., Urban, J.: Deepmath - deep
  sequence models for premise selection. CoRR  \textbf{abs/1606.04442} (2016)

\bibitem{DBLP:journals/corr/abs-2006-11259}
Ayg{\"{u}}n, E., Ahmed, Z., Anand, A., Firoiu, V., Glorot, X., Orseau, L.,
  Precup, D., Mourad, S.: Learning to prove from synthetic theorems. CoRR
  \textbf{abs/2006.11259} (2020)

\bibitem{DBLP:books/el/RV01/BachmairG01}
Bachmair, L., Ganzinger, H.: Resolution theorem proving. In: Robinson, J.A.,
  Voronkov, A. (eds.) Handbook of Automated Reasoning (in 2 volumes), pp.
  19--99. Elsevier and {MIT} Press (2001).
  \doi{10.1016/b978-044450813-3/50004-7}

\bibitem{BarFT-SMTLIB}
Barrett, C., Fontaine, P., Tinelli, C.: {The Satisfiability Modulo Theories
  Library (SMT-LIB)}. \url{www.SMT-LIB.org} (2016)

\bibitem{DBLP:journals/jfrea/BlanchetteKPU16}
Blanchette, J.C., Kaliszyk, C., Paulson, L.C., Urban, J.: Hammering towards
  {QED}. J. Formaliz. Reason.  \textbf{9}(1),  101--148 (2016).
  \doi{10.6092/issn.1972-5787/4593}

\bibitem{DBLP:conf/cade/ChvalovskyJ0U19}
Chvalovsk{\'{y}}, K., Jakubuv, J., Suda, M., Urban, J.: {ENIGMA-NG:} efficient
  neural and gradient-boosted inference guidance for {E}. In: Fontaine, P.
  (ed.) Automated Deduction - {CADE} 27 - 27th International Conference on
  Automated Deduction, Natal, Brazil, August 27-30, 2019, Proceedings. LNCS,
  vol. 11716, pp. 197--215. Springer (2019).
  \doi{10.1007/978-3-030-29436-6\_12}

\bibitem{DBLP:journals/corr/abs-1911-02065}
Crouse, M., Whitehead, S., Abdelaziz, I., Makni, B., Cornelio, C., Kapanipathi,
  P., Pell, E., Srinivas, K., Thost, V., Witbrock, M., Fokoue, A.: A deep
  reinforcement learning based approach to learning transferable proof guidance
  strategies. CoRR  \textbf{abs/1911.02065} (2019)

\bibitem{DS1996b}
Denzinger, J., Schulz, S.: {Learning Domain Knowledge to Improve Theorem
  Proving}. In: McRobbie, M., Slaney, J. (eds.) Proc.\ of the 13th CADE, New
  Brunswick. pp. 62--76. No.~1104 in LNAI, Springer (1996)

\bibitem{DBLP:conf/frocos/FarberK15}
F{\"{a}}rber, M., Kaliszyk, C.: Random forests for premise selection. In: Lutz,
  C., Ranise, S. (eds.) Frontiers of Combining Systems - 10th International
  Symposium, FroCoS 2015, Wroclaw, Poland, September 21-24, 2015. Proceedings.
  LNCS, vol.~9322, pp. 325--340. Springer (2015).
  \doi{10.1007/978-3-319-24246-0\_20}

\bibitem{DBLP:conf/cade/FarberKU17}
F{\"{a}}rber, M., Kaliszyk, C., Urban, J.: Monte carlo tableau proof search.
  In: de~Moura, L. (ed.) Automated Deduction - {CADE} 26 - 26th International
  Conference on Automated Deduction, Gothenburg, Sweden, August 6-11, 2017,
  Proceedings. LNCS, vol. 10395, pp. 563--579. Springer (2017).
  \doi{10.1007/978-3-319-63046-5\_34}

\bibitem{DBLP:conf/ijcai/Gleiss020}
Gleiss, B., Suda, M.: Layered clause selection for saturation-based theorem
  proving. In: Fontaine, P., Korovin, K., Kotsireas, I.S., R{\"{u}}mmer, P.,
  Tourret, S. (eds.) Joint Proceedings of the 7th Workshop on Practical Aspects
  of Automated Reasoning {(PAAR)} and the 5th Satisfiability Checking and
  Symbolic Computation Workshop (SC-Square), co-located with the 10th
  International Joint Conference on Automated Reasoning {(IJCAR} 2020), Paris,
  France, June-July, 2020 (Virtual). {CEUR} Workshop Proceedings, vol.~2752,
  pp. 34--52. CEUR-WS.org (2020)

\bibitem{DBLP:conf/cade/Gleiss020}
Gleiss, B., Suda, M.: Layered clause selection for theory reasoning - (short
  paper). In: Peltier, N., Sofronie{-}Stokkermans, V. (eds.) Automated
  Reasoning - 10th International Joint Conference, {IJCAR} 2020, Paris, France,
  July 1-4, 2020, Proceedings, Part {I}. LNCS, vol. 12166, pp. 402--409.
  Springer (2020). \doi{10.1007/978-3-030-51074-9\_23}

\bibitem{DBLP:conf/icnn/GollerK96}
Goller, C., K{\"{u}}chler, A.: Learning task-dependent distributed
  representations by backpropagation through structure. In: Proceedings of
  International Conference on Neural Networks (ICNN'96), Washington, DC, USA,
  June 3-6, 1996. pp. 347--352. {IEEE} (1996). \doi{10.1109/ICNN.1996.548916}

\bibitem{DBLP:books/daglib/0040158}
Goodfellow, I.J., Bengio, Y., Courville, A.C.: Deep Learning. Adaptive
  computation and machine learning, {MIT} Press (2016)

\bibitem{DBLP:journals/jfrea/GrabowskiKN10}
Grabowski, A., Kornilowicz, A., Naumowicz, A.: Mizar in a nutshell. J.
  Formaliz. Reason.  \textbf{3}(2),  153--245 (2010).
  \doi{10.6092/issn.1972-5787/1980}

\bibitem{DBLP:conf/cade/HoderV11}
Hoder, K., Voronkov, A.: Sine qua non for large theory reasoning. In:
  Bj{\o}rner, N., Sofronie{-}Stokkermans, V. (eds.) 23rd International
  Conference on Automated Deduction (CADE 2011). LNCS, vol.~6803, pp. 299--314.
  Springer (2011). \doi{10.1007/978-3-642-22438-6\_23}

\bibitem{DBLP:conf/cade/JakubuvCOP0U20}
Jakubuv, J., Chvalovsk{\'{y}}, K., Ols{\'{a}}k, M., Piotrowski, B., Suda, M.,
  Urban, J.: {ENIGMA} anonymous: Symbol-independent inference guiding machine
  (system description). In: Peltier, N., Sofronie{-}Stokkermans, V. (eds.)
  Automated Reasoning - 10th International Joint Conference, {IJCAR} 2020,
  Paris, France, July 1-4, 2020, Proceedings, Part {II}. LNCS, vol. 12167, pp.
  448--463. Springer (2020). \doi{10.1007/978-3-030-51054-1\_29}

\bibitem{DBLP:conf/mkm/JakubuvU17}
Jakubuv, J., Urban, J.: {ENIGMA:} efficient learning-based inference guiding
  machine. In: Geuvers, H., England, M., Hasan, O., Rabe, F., Teschke, O.
  (eds.) Intelligent Computer Mathematics - 10th International Conference,
  {CICM} 2017, Edinburgh, UK, July 17-21, 2017, Proceedings. LNCS, vol. 10383,
  pp. 292--302. Springer (2017). \doi{10.1007/978-3-319-62075-6\_20}

\bibitem{DBLP:conf/mkm/JakubuvU18}
Jakubuv, J., Urban, J.: Enhancing {ENIGMA} given clause guidance. In: Rabe, F.,
  Farmer, W.M., Passmore, G.O., Youssef, A. (eds.) Intelligent Computer
  Mathematics - 11th International Conference, {CICM} 2018, Hagenberg, Austria,
  August 13-17, 2018, Proceedings. LNCS, vol. 11006, pp. 118--124. Springer
  (2018). \doi{10.1007/978-3-319-96812-4\_11}

\bibitem{DBLP:conf/itp/JakubuvU19}
Jakubuv, J., Urban, J.: Hammering {Mizar} by learning clause guidance (short
  paper). In: Harrison, J., O'Leary, J., Tolmach, A. (eds.) 10th International
  Conference on Interactive Theorem Proving, {ITP} 2019, September 9-12, 2019,
  Portland, OR, {USA}. LIPIcs, vol.~141, pp. 34:1--34:8. Schloss Dagstuhl -
  Leibniz-Zentrum f{\"{u}}r Informatik (2019). \doi{10.4230/LIPIcs.ITP.2019.34}

\bibitem{DBLP:journals/jar/KaliszykU15a}
Kaliszyk, C., Urban, J.: Mizar 40 for mizar 40. J. Autom. Reason.
  \textbf{55}(3),  245--256 (2015). \doi{10.1007/s10817-015-9330-8}

\bibitem{DBLP:conf/cav/KovacsV13}
Kov{\'{a}}cs, L., Voronkov, A.: First-order theorem proving and {Vampire}. In:
  Sharygina, N., Veith, H. (eds.) Computer Aided Verification - 25th
  International Conference, {CAV} 2013, Saint Petersburg, Russia, July 13-19,
  2013. Proceedings. LNCS, vol.~8044, pp. 1--35. Springer (2013).
  \doi{10.1007/978-3-642-39799-8\_1}

\bibitem{DBLP:conf/iclr/LooksHHN17}
Looks, M., Herreshoff, M., Hutchins, D., Norvig, P.: Deep learning with dynamic
  computation graphs. In: 5th International Conference on Learning
  Representations, {ICLR} 2017, Toulon, France, April 24-26, 2017, Conference
  Track Proceedings. OpenReview.net (2017)

\bibitem{DBLP:conf/lpar/LoosISK17}
Loos, S.M., Irving, G., Szegedy, C., Kaliszyk, C.: Deep network guided proof
  search. In: Eiter, T., Sands, D. (eds.) LPAR-21, 21st International
  Conference on Logic for Programming, Artificial Intelligence and Reasoning,
  Maun, Botswana, May 7-12, 2017. EPiC Series in Computing, vol.~46, pp.
  85--105. EasyChair (2017)

\bibitem{DBLP:books/el/RV01/NieuwenhuisR01}
Nieuwenhuis, R., Rubio, A.: Paramodulation-based theorem proving. In: Robinson,
  J.A., Voronkov, A. (eds.) Handbook of Automated Reasoning (in 2 volumes), pp.
  371--443. Elsevier and {MIT} Press (2001).
  \doi{10.1016/b978-044450813-3/50009-6}

\bibitem{NEURIPS2019_9015}
Paszke, A., Gross, S., Massa, F., Lerer, A., Bradbury, J., Chanan, G., et~al.:
  Pytorch: An imperative style, high-performance deep learning library. In:
  Wallach, H., Larochelle, H., Beygelzimer, A., d\textquotesingle
  Alch\'{e}-Buc, F., Fox, E., Garnett, R. (eds.) Advances in Neural Information
  Processing Systems 32, pp. 8024--8035. Curran Associates, Inc. (2019),
  \url{http://papers.neurips.cc/paper/9015-pytorch-an-imperative-style-high-performance-deep-learning-library.pdf}

\bibitem{DBLP:conf/lpar/PiotrowskiU20}
Piotrowski, B., Urban, J.: Stateful premise selection by recurrent neural
  networks. In: Albert, E., Kov{\'{a}}cs, L. (eds.) {LPAR} 2020: 23rd
  International Conference on Logic for Programming, Artificial Intelligence
  and Reasoning, Alicante, Spain, May 22-27, 2020. EPiC Series in Computing,
  vol.~73, pp. 409--422. EasyChair (2020),
  \url{https://easychair.org/publications/paper/g38n}

\bibitem{NIPS2011_218a0aef}
Recht, B., Re, C., Wright, S., Niu, F.: Hogwild!: A lock-free approach to
  parallelizing stochastic gradient descent. In: Shawe-Taylor, J., Zemel, R.,
  Bartlett, P., Pereira, F., Weinberger, K.Q. (eds.) Advances in Neural
  Information Processing Systems. vol.~24. Curran Associates, Inc. (2011),
  \url{https://proceedings.neurips.cc/paper/2011/file/218a0aefd1d1a4be65601cc6ddc1520e-Paper.pdf}

\bibitem{DBLP:conf/cade/RegerSV15}
Reger, G., Suda, M., Voronkov, A.: Playing with {AVATAR}. In: Felty, A.P.,
  Middeldorp, A. (eds.) Automated Deduction - {CADE-25} - 25th International
  Conference on Automated Deduction, Berlin, Germany, August 1-7, 2015,
  Proceedings. LNCS, vol.~9195, pp. 399--415. Springer (2015).
  \doi{10.1007/978-3-319-21401-6\_28}

\bibitem{DBLP:journals/jsc/RiazanovV03}
Riazanov, A., Voronkov, A.: Limited resource strategy in resolution theorem
  proving. J. Symb. Comput.  \textbf{36}(1-2),  101--115 (2003).
  \doi{10.1016/S0747-7171(03)00040-3}

\bibitem{Schulz:Diss-2000}
Schulz, S.: {Learning Search Control Knowledge for Equational Deduction}.
  No.~230 in DISKI, Akademische Verlagsgesellschaft Aka GmbH Berlin (2000)

\bibitem{SCV:CADE-2019}
Schulz, S., Cruanes, S., Vukmirovic, P.: Faster, higher, stronger: {E} 2.3. In:
  Fontaine, P. (ed.) Automated Deduction - {CADE} 27 - 27th International
  Conference on Automated Deduction, Natal, Brazil, August 27-30, 2019,
  Proceedings. LNCS, vol. 11716, pp. 495--507. Springer (2019).
  \doi{10.1007/978-3-030-29436-6\_29}

\bibitem{DBLP:conf/cade/SchulzM16}
Schulz, S., M{\"{o}}hrmann, M.: Performance of clause selection heuristics for
  saturation-based theorem proving. In: Olivetti, N., Tiwari, A. (eds.)
  Automated Reasoning - 8th International Joint Conference, {IJCAR} 2016,
  Coimbra, Portugal, June 27 - July 2, 2016, Proceedings. LNCS, vol.~9706, pp.
  330--345. Springer (2016). \doi{10.1007/978-3-319-40229-1\_23}

\bibitem{DBLP:journals/jmlr/SrivastavaHKSS14}
Srivastava, N., Hinton, G.E., Krizhevsky, A., Sutskever, I., Salakhutdinov, R.:
  Dropout: a simple way to prevent neural networks from overfitting. J. Mach.
  Learn. Res.  \textbf{15}(1),  1929--1958 (2014),
  \url{http://dl.acm.org/citation.cfm?id=2670313}

\bibitem{Vampire2019:Aiming_for_Goal_with}
Suda, M.: Aiming for the goal with {SInE}. In: Kov{\'{a}}cs, L., Voronkov, A.
  (eds.) Vampire 2018 and Vampire 2019. The 5th and 6th Vampire Workshops. EPiC
  Series in Computing, vol.~71, pp. 38--44. EasyChair (2020).
  \doi{10.29007/q4pt}

\bibitem{SudaCADE2021}
Suda, M.: Improving {ENIGMA}-style clause selection while learning from
  history. In: Platzer, A., Sutcliffe, G. (eds.) Proceedings of the 28th CADE
  (2021), to appear. See also \url{https://arxiv.org/abs/2102.13564}

\bibitem{DBLP:conf/cade/Tammet19}
Tammet, T.: {GKC:} {A} reasoning system for large knowledge bases. In:
  Fontaine, P. (ed.) Automated Deduction - {CADE} 27 - 27th International
  Conference on Automated Deduction, Natal, Brazil, August 27-30, 2019,
  Proceedings. LNCS, vol. 11716, pp. 538--549. Springer (2019).
  \doi{10.1007/978-3-030-29436-6\_32}

\bibitem{DBLP:journals/jar/Urban06}
Urban, J.: {MPTP} 0.2: Design, implementation, and initial experiments. J.
  Autom. Reason.  \textbf{37}(1-2),  21--43 (2006).
  \doi{10.1007/s10817-006-9032-3}

\bibitem{DBLP:conf/tableaux/UrbanVS11}
Urban, J., Vyskocil, J., Step{\'{a}}nek, P.: Malecop machine learning
  connection prover. In: Br{\"{u}}nnler, K., Metcalfe, G. (eds.) Automated
  Reasoning with Analytic Tableaux and Related Methods - 20th International
  Conference, {TABLEAUX} 2011, Bern, Switzerland, July 4-8, 2011. Proceedings.
  LNCS, vol.~6793, pp. 263--277. Springer (2011).
  \doi{10.1007/978-3-642-22119-4\_21}

\bibitem{DBLP:conf/cav/Voronkov14}
Voronkov, A.: {AVATAR:} the architecture for first-order theorem provers. In:
  Biere, A., Bloem, R. (eds.) Computer Aided Verification - 26th International
  Conference, {CAV} 2014, Held as Part of the Vienna Summer of Logic, {VSL}
  2014, Vienna, Austria, July 18-22, 2014. Proceedings. LNCS, vol.~8559, pp.
  696--710. Springer (2014). \doi{10.1007/978-3-319-08867-9\_46}

\bibitem{DBLP:conf/nips/WangTWD17}
Wang, M., Tang, Y., Wang, J., Deng, J.: Premise selection for theorem proving
  by deep graph embedding. In: Guyon, I., von Luxburg, U., Bengio, S., Wallach,
  H.M., Fergus, R., Vishwanathan, S.V.N., Garnett, R. (eds.) Advances in Neural
  Information Processing Systems 30: Annual Conference on Neural Information
  Processing Systems 2017, December 4-9, 2017, Long Beach, CA, {USA}. pp.
  2786--2796 (2017),
  \url{https://proceedings.neurips.cc/paper/2017/hash/18d10dc6e666eab6de9215ae5b3d54df-Abstract.html}

\bibitem{DBLP:conf/cade/WeidenbachDFKSW09}
Weidenbach, C., Dimova, D., Fietzke, A., Kumar, R., Suda, M., Wischnewski, P.:
  {SPASS} version 3.5. In: Schmidt, R.A. (ed.) Automated Deduction - CADE-22,
  22nd International Conference on Automated Deduction, Montreal, Canada,
  August 2-7, 2009. Proceedings. LNCS, vol.~5663, pp. 140--145. Springer
  (2009). \doi{10.1007/978-3-642-02959-2\_10}

\end{thebibliography}

\end{document}